\documentclass[conference]{IEEEtran}
\IEEEoverridecommandlockouts

\usepackage{amsmath,amssymb,amsfonts}
\usepackage{algorithmic}
\usepackage{graphicx}
\usepackage{textcomp}
\usepackage{xcolor}
\usepackage{siunitx}
\usepackage{xcolor}
\usepackage{siunitx}
\usepackage{hyperref}
\usepackage{siunitx}
\usepackage{caption}
\usepackage[list=true]{subcaption}

\usepackage{xcolor}

\begin{document}

\title{Exploitation of Hidden Context in Dynamic Movement Forecasting: A Neural Network Journey from Recurrent to Graph Neural Networks and General Purpose Transformers\\
{\footnotesize \thanks{This work has been carried out within the DARCII project, funding code 50NA2401, sponsored by the German Federal Ministry for Economic Affairs and Climate Action (BMWK) and supported by the German Aerospace Center (DLR), the Bundesnetzagentur (BNetzA), and the Federal Agency for Cartography and Geodesy (BKG). This work was also supported by the Bavarian Ministry for Economic Affairs, Infrastructure, Transport and Technology through the Center for Analytics Data Applications (ADA-Center) within the framework of ``BAYERN DIGITAL II'' (20-3410-2-9-8).}}}

\author{\IEEEauthorblockN{Lukas Schelenz\IEEEauthorrefmark{1},
    Shobha Rajanna\IEEEauthorrefmark{1},
    Denis Gosalci\IEEEauthorrefmark{1},
    Lucas Heublein\IEEEauthorrefmark{1},
    Jonas Pirkl\IEEEauthorrefmark{1},
    Jonathan Ott\IEEEauthorrefmark{1},\\
    Felix Ott\IEEEauthorrefmark{1},
    Christopher Mutschler\IEEEauthorrefmark{1},
    Tobias Feigl\IEEEauthorrefmark{1}\IEEEauthorrefmark{2}
  }
  \IEEEauthorblockA{\IEEEauthorrefmark{1}Fraunhofer Institute for Integrated Circuits IIS, 90411 Nürnberg, Germany}
  \IEEEauthorblockA{\IEEEauthorrefmark{2}Friedrich-Alexander-Universität Erlangen-Nürnberg, 91058 Erlangen, Germany}
  \IEEEauthorblockA{\{lukas.schelenz, shobha.rajanna, denis.gosalci, lucas.heublein, jonas.pirkl, jonathan.ott,\\felix.ott, christopher.mutschler, tobias.feigl\}@iis.fraunhofer.de}
}
\maketitle
\begin{abstract}
Forecasting within signal processing pipelines is crucial for mitigating delays, particularly in predicting the dynamic movements of objects such as NBA players. This task poses significant challenges due to the inherently interactive and unpredictable nature of sports, where abrupt changes in velocity and direction are prevalent. Traditional approaches, including (S)ARIMA(X), Kalman filters (KF), and Particle filters (PF), often struggle to model the non-linear dynamics present in such scenarios. Machine learning (ML) methods, such as long short-term memory (LSTM) networks, graph neural networks (GNNs), and Transformers, offer greater flexibility and accuracy but frequently fail to explicitly capture the interplay between temporal dependencies and contextual interactions, which are critical in chaotic sports environments. 

In this paper, we evaluate these models and assess their strengths and weaknesses. Experimental results reveal key performance trade-offs across input history length, generalizability, and the ability to incorporate contextual information. ML-based methods demonstrated substantial improvements over linear models across forecast horizons of up to 2\unit{s}. Among the tested architectures, our hybrid LSTM augmented with contextual information achieved the lowest final displacement error (FDE) of 1.51\unit{m}, outperforming temporal convolutional neural network (TCNN), graph attention network (GAT), and Transformers, while also requiring less data and training time compared to GAT and Transformers. Our findings indicate that no single architecture excels across all metrics, emphasizing the need for task-specific considerations in trajectory prediction for fast-paced, dynamic environments such as NBA gameplay.
\end{abstract}

\section{Introduction}

In recent years, ML-driven forecasting has gained significant attention across various domains, including sports analytics~\cite{brefeld2017guest}, stock market predictions~\cite{rapach2020,siam_namini_namin}, handwriting classification~\cite{ott_wacv}, electricity pricing analytics~\cite{ribeiro2021}, weather forecasting~\cite{mills2019}, robotic trajectory prediction~\cite{ott_cvprw,ott_fusing}, and healthcare. Within the realm of sports analytics, research has predominantly concentrated on predicting game outcomes~\cite{morgulev2018sports,lam2018one}, while relatively few studies have addressed the forecasting of human motion, particularly the trajectories of athletes~\cite{feigl2021datengetriebene}. Accurate trajectory forecasting plays a pivotal role in enhancing real-time game analysis, offering tactical advantages~\cite{wei2022rethinking}, and mitigating injury risks~\cite{zhang1907stochastic,e23040477}.

Athlete movements are inherently nonlinear and dynamic, influenced by factors such as spatial positioning, proximity to opponents, and abrupt changes in speed and direction~\cite{moreObjectsMakeSence}. Traditional statistical models, including ARIMA~\cite{siam_namini_namin} and linear regression, as well as Bayesian methods such as KFs~\cite{kalman} and PFs~\cite{Particle}, rely on linearity assumptions that render them inadequate for capturing sudden transitions and complex multi-agent interactions~\cite{feigl2021datengetriebene,10.1145/3409334.3452064}. These limitations restrict their ability to effectively handle long-term dependencies and contextual interactions in dynamic environments. To address these challenges, ML-based models have been explored for their capability to learn nonlinear patterns. Recurrent neural networks (RNNs), particularly LSTM networks~\cite{LSTM}, gated recurrent units (GRUs)~\cite{GRU}, and Legendre memory units (LMUs)~\cite{lmu}, are effective in capturing sequential patterns and temporal dependencies. However, they may suffer from vanishing memory and limited scalability when processing extended sequences~\cite{transformer}. GNNs~\cite{GNN,GNNmultivariate,xu2022adaptive,INTROGNN,traficgnn}, particularly GATs~\cite{mo2021graph,GAT}, offer improved modeling of multi-agent interactions and provide a more interpretable framework for interaction analysis. Nonetheless, GNNs often lack temporal depth, reducing their effectiveness in sequential tasks. Transformers~\cite{transformer,Transformers,bert,carion2020,dosovitskiy2020}, originally developed for natural language processing (NLP), present an alternative approach by employing self-attention mechanisms to capture long-range dependencies and global patterns. Despite their demonstrated success in various fields, these architectures remain underexplored for human motion forecasting in sports~\cite{alahi2016social,mohamed2020social}. Their adoption has been constrained by high computational costs and the absence of explicit domain-specific contextual modeling, which limits their applicability to real-time sports analytics.

\textbf{Contributions.} This study presents a comprehensive evaluation and comparative analysis of several ML-based forecasting models, including TCNNs, LSTMs, LMUs, GATs, and Transformers, and their hybrid configurations, for predicting the short-term movements of players in fast-paced and dynamic environments exemplified by NBA games. The investigation focuses on quantifying performance trade-offs across multiple criteria, including input history length, computational complexity, generalization capacity, and the ability to leverage contextual interactions. To address limitations in existing approaches, the proposed framework integrates explicit contextual modeling within a recurrent architecture. Unlike conventional methods that treat spatial and temporal features independently, this approach incorporates transformed distance functions within an LSTM pipeline, emphasizing the representation of shorter, more informative proximities. Additionally, team dynamics and inter-player interactions are explicitly modeled through hybrid architectures, such as RNN-GAT-RNN and fusion models that combine recurrent and attention-based mechanisms. The inclusion of Transformer-based components enables the framework to effectively capture both global dependencies and fine-grained dynamics, overcoming scalability challenges through domain-specific encodings. These innovations collectively enable the models to outperform state-of-the-art approaches in trajectory prediction, particularly in chaotic environments characterized by abrupt, non-linear player movements and complex inter-player interactions, such as those observed in NBA basketball games.

\textbf{Outline.} Sec.~\ref{sec:related_work} offers a review of related literature. Sec.~\ref{sec:methodology} describes preprocessing pipeline and the proposed methods. Sec.~\ref{sec:experiments} details the experimental setup, followed by the discussion of results in Sec.~\ref{sec:results}. Sec.~\ref{sec:summary} concludes the paper.
\section{Related Work}
\label{sec:related_work}

\textbf{Classical Approaches.} Human motion forecasting has applications across a wide range of domains, including public safety~\cite{mo2021graph}, healthcare~\cite{brefeld2017guest}, and sports analytics~\cite{li2019grip++,feigl2021datengetriebene,morgulev2018sports,lam2018one}. In the context of basketball, accurate player trajectory prediction offers substantial benefits for tactical decision-making, performance evaluation, and real-time analytics. Traditionally, model-based approaches such as KF~\cite{kalman} and PF~\cite{Particle} have been employed to address this task. These methods are effective in scenarios where motion is linear or near-linear, leveraging recent observations to generate predictions with computational efficiency. However, their performance is significantly constrained in dynamic, multi-agent environments such as basketball, where player movements are characterized by abrupt changes in velocity and direction. The reliance on simplistic motion assumptions and limited historical context further diminishes their effectiveness in capturing the complexities of real-world trajectories~\cite{feigl2021datengetriebene}.

\textbf{Spatial and Temporal Models.} ML-based approaches have increasingly supplanted traditional statistical methods, offering greater flexibility in modeling complex, non-linear patterns. CNNs~\cite{CNN_goodfellow,CNN3,cnn_emre}, for example, are adept at extracting spatial features from trajectory data but often fall short in capturing temporal dependencies and inter-player dynamics. Nikhil et al.~\cite{CNN_SRC} demonstrated the use of CNNs for player position prediction. However, the model's inability to represent complex interactions or long-term temporal dependencies limited its applicability to team sports. RNNs, particularly LSTMs~\cite{LSTM2}, address these limitations by modeling sequential data and leveraging historical information to capture long-term dependencies. This capability has led to notable performance gains over KFs and PFs in dynamic environments~\cite{feigl2021datengetriebene}. Nonetheless, the reliance of LSTMs on gating mechanisms to control memory introduces computational inefficiencies. LMUs~\cite{lmu} seek to mitigate these challenges by employing a mathematically grounded state-space approach, enabling the effective retention of long-term dependencies while minimizing the dependence on learned gating mechanisms.

\textbf{Transformer Architectures.} Originally developed for NLP, Transformers~\cite{transformer} extend the capabilities of sequence modeling by employing self-attention mechanisms. These architectures effectively capture long-range dependencies while mitigating the vanishing gradient problems typically encountered in RNNs. Recent research has demonstrated the potential of Transformers in motion prediction tasks. For instance, Giuliari et al.~\cite{rel_work_denis} applied Transformers to static motion data, while Mascaro et al.~\cite{human_transformer} introduced a framework integrating spatial and temporal attention for 3D skeletal motion prediction. Transformers excel at simultaneously capturing global dependencies and fine-grained local patterns, theoretically making them well-suited for complex trajectory prediction problems. However, their practical application in real-time sports contexts remains limited due to high computational demands and the absence of explicit interaction modeling, that hinder their ability to effectively capture dynamic inter-agent relationships.

\textbf{Graph Neural Networks.} There are many adaptations of GNNs~\cite{mo2021graph,mo2020interaction,almahmoud2023holistic} that enable interaction-aware modeling. By representing players as nodes and interactions, such as spatial proximity or passes, as edges, GNNs provide a powerful framework for capturing inter-player dynamics. Mo et al.~\cite{mo2021graph} demonstrated the effectiveness of GNNs in vehicle trajectory forecasting, while Xu et al.~\cite{xu2022adaptive} advanced this approach by introducing attention-based GNNs (GATs) to model adaptive interaction strengths. Despite their demonstrated utility in other domains, these methods have not been adapted for basketball, where the complexity of team strategies demands a nuanced understanding of both spatial and temporal relationships. Effective modeling this context requires capturing dynamic interactions and evolving game scenarios, which remain unexplored using existing GNN-based approaches.

\textbf{Summary.} Existing literature does not sufficiently address the complex interplay between temporal dependencies, inter-player interactions, and the dynamic context inherent in sports environments. Traditional statistical models are overly simplistic, relying on assumptions that limit their applicability in capturing the nuanced, non-linear motion patterns characteristic of such settings. ML-based approaches, while more flexible, frequently prioritize either temporal or spatial aspects, with limited efforts to seamlessly integrate both dimensions. The reviewed methods fail to incorporate context information, resulting in suboptimal performance for forecasting dynamic motion in sports. These limitations highlight the need for methods that effectively bridge these gaps, enabling more accurate and context-aware trajectory predictions.
\section{Methodology}
\label{sec:methodology}

This Section presents the general and model-specific preprocessing pipelines (Sec.~\ref{sec:preproc}). Sec.~\ref{sec:mainproc} introduces all methods. Sec.~\ref{postprocessing} explains method-specific postprocessing.

\subsection{Preprocessing}\label{sec:preproc}

\begin{figure}[!t]
    \centering
    \includegraphics[width=1\linewidth]{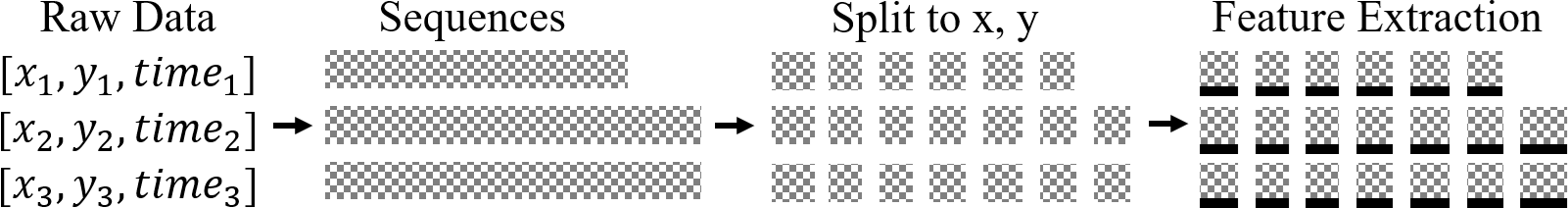}
    \caption{Preprocessing of the NBA dataset.}
    \label{fig:dataloader_pipeline_lukas}
\end{figure}

\textbf{General.} To effectively train the models, several preprocessing steps are applied, see Fig.~\ref{fig:dataloader_pipeline_lukas}. Initially, raw data is collected and transformed into a unified structure. Each trajectory is represented as a 3D tensor of shape \text{[num. timesteps, num. objects, num. features]}, where \text{num. timesteps} refers to the trajectory duration, \text{num. objects} represents the entities in the scene, e.g., players, ball, goals, and \text{num. features} includes attributes such as absolute position ($pos_x$, $pos_y$) and velocity ($v_x$, $v_y$). The NBA dataset is sampled at 25\,\unit{Hz}. Interpolation is applied to all trajectories to ensure a uniform time interval of 0.04\,\unit{s}. The dataset is then split into training, validation, and test sets, with 90\% of the data allocated for training, further divided into 90\% for training and 10\% for validation. The test set is sourced from a single, unseen game to maintain evaluation integrity. To enhance generalization, trajectories in the training set are shuffled, while those in the test set remain unshuffled to preserve temporal continuity. 
Finally, the data is segmented into overlapping windows using a sliding window approach (step size = 1 to capture all windows).
Each window is defined by a \text{window size} comprising a \text{history length} (input context) and a \text{forecast length} (steps to predict). 
For each window we convert its positions to velocities ($v_x$, $v_y$). 
All these position and velocity features are normalized. 
So, we ensure consistent magnitudes and prevent large positional values from dominating the training process. 

\vspace{0.1cm}
\textbf{Constant Velocity, Linear, TCNN, LSTM and LMU} may directly use features (\textit{positions} and / or \textit{velocities}) of the general preprocessing pipeline. 

\vspace{0.1cm}
\textbf{GNN} builds upon the general preprocessing pipeline with additional steps to transform the data into a graph representation suitable for graph-based learning. For each trajectory segment, a graph is constructed, where nodes represent entities such as players, and edges capture interactions or relationships, such as spatial proximity or passing connections. Each node is assigned a feature vector containing attributes such as \textit{positional} coordinates ($x$, $y$), \textit{velocity} components ($v_x$, $v_y$), and other contextual features, such as distances to opponents or team affiliation. The graph's connectivity is encoded using an edge index tensor, which is stored alongside the node features. These graph representations preserve both individual characteristics and relational dynamics, enabling GNN to learn from the temporal interactions within the scene. The dataset is then organized into batches, with each batch consisting of graphs corresponding to individual trajectory segments.

\vspace{0.1cm}
\textbf{Transformer} follows the general pipeline with adjustments specific to the architecture. Positional encoding is applied to the input data to explicitly capture temporal dependencies, enhancing the input tensor (\textit{position}, \textit{velocity}) and enabling the attention mechanism to focus on key timesteps. These features are normalized and scaled to ensure relevance for the attention mechanism. The resulting tensors, which combine normalized trajectory features and positional encodings, enable the model to effectively leverage both spatial and temporal relationships for prediction tasks. For the Transformer model specifically, a CNN-based temporal embedding is used to implicitly encode local sequential patterns before the data enters the attention layers, enhancing the representation capacity without relying on explicit positional encodings. This is inspired by Baevski et al.~\cite{baevski2020}, that use convolutional layers to extract temporal features directly from raw sequences, allowing the model to learn local time-dependent structures more effectively.

\subsection{Main Processing}\label{sec:mainproc}

{In preliminary studies, we optimized each model individually by varying combinations of input features, including positions, velocities, distances, and context information.\footnote{{Note that, if a model only predicts velocities, we reconstruct positions by concatenating them starting from the last known, i.e., initial, position.\label{fn_initialpos}}} Here, we report only the best-performing configurations for each model w.r.t. a fixed input sequence length of 2\unit{s}.\footnote{{In individual tests, we found that no model benefits from an input sequence length longer than 2\unit{s}, see our discussion in Sec.~\ref{sec:results}.}}

\vspace{0.1cm}
\textbf{Constant Velocity} naively extrapolates future positions by taking the \textit{velocity} from the last known position and applying it uniformly across all forecast steps.

\vspace{0.1cm}
{\textbf{Linear} may capture temporal context dependencies as our flattened sequential input (\textit{velocities}) provides it. Its architecture includes 2 fully connected layers (128 neurons each), followed by ReLU functions and dropout for regularization.}}

\vspace{0.1cm}
{\textbf{TCNN} uses 1D causal convolutions to ensure the output at time t only depends on inputs (\textit{velocities}) at or before t~\cite{lea2017temporal}. We found 5 convolutional layers with increasing dilation factors (1, 2, 4, 8) capture long-range temporal dependencies efficiently. Each layer uses residual connections, weight normalization, and ReLU activations. A Kernel size of 2 and 42 filters control its capacity and receptive field. TCNNs process in parallel, train faster, and maintain temporal causality.}

\vspace{0.1cm}
\textbf{LSTM.} Memory cell-based models, such as LSTMs~\cite{LSTM}, are particularly effective at capturing complex temporal dependencies in sequential data. This architecture allows for the storage and selective processing of information across multiple timesteps, enabling the model to identify and leverage even distant dependencies within the data. In scenarios where past events significantly influence future outcomes, the ability to learn long-term dependencies is crucial. Capturing these relationships facilitates more accurate predictions and trajectory calculations. The operation of an LSTM cell relies on the interaction of several components that regulate the flow of information, including the input gate, forgetting gate, output gate, and cell state. Each of these gates plays a specific role in the process of information storage and processing. 

The mathematical description of these cell operations is:
\begin{equation}
\begin{split}
        i_t &= \sigma(W_{ii}x_t+b_{ii} + W_{hi}h_{t-1}+b_{hi}), \\
        f_t &= \sigma(W_{if}x_t+b_{if} + W_{hf}h_{t-1}+b_{hf}), \\
        g_t &= tanh(W_{ig}x_t+b_{ig} + W_{hg}h_{t-1}+b_{hg}), \\
        o_t &= \sigma(W_{io}x_t+b_{io} + W_{ho}h_{t-1}+b_{ho}), \\
        c_t &= f_{t} \odot c{t-1}+i_{t}\odot g_t, \\
        h_t &= o_{t} \odot tanh(c_t),
\end{split}
\label{lstm_equation}
\end{equation}
where $h_t$ is the hidden state at time $t$, $c_t$ is the cell state at time $t$, $x_t$ is the input at time $t$, $h_{t-1}$ is the hidden state of the layer at time $t-1$ or the initial hidden state at time $0$, and $i_t$, $f_t$, $g_t$, and $o_t$ are the input, forgotten, hidden, and output gates, with sigmoid function $\sigma$ and Hadamard product $\odot$. 

\begin{figure}[!t]
    \centering
    \includegraphics[width=1\linewidth,trim={0 120 0 0},clip]{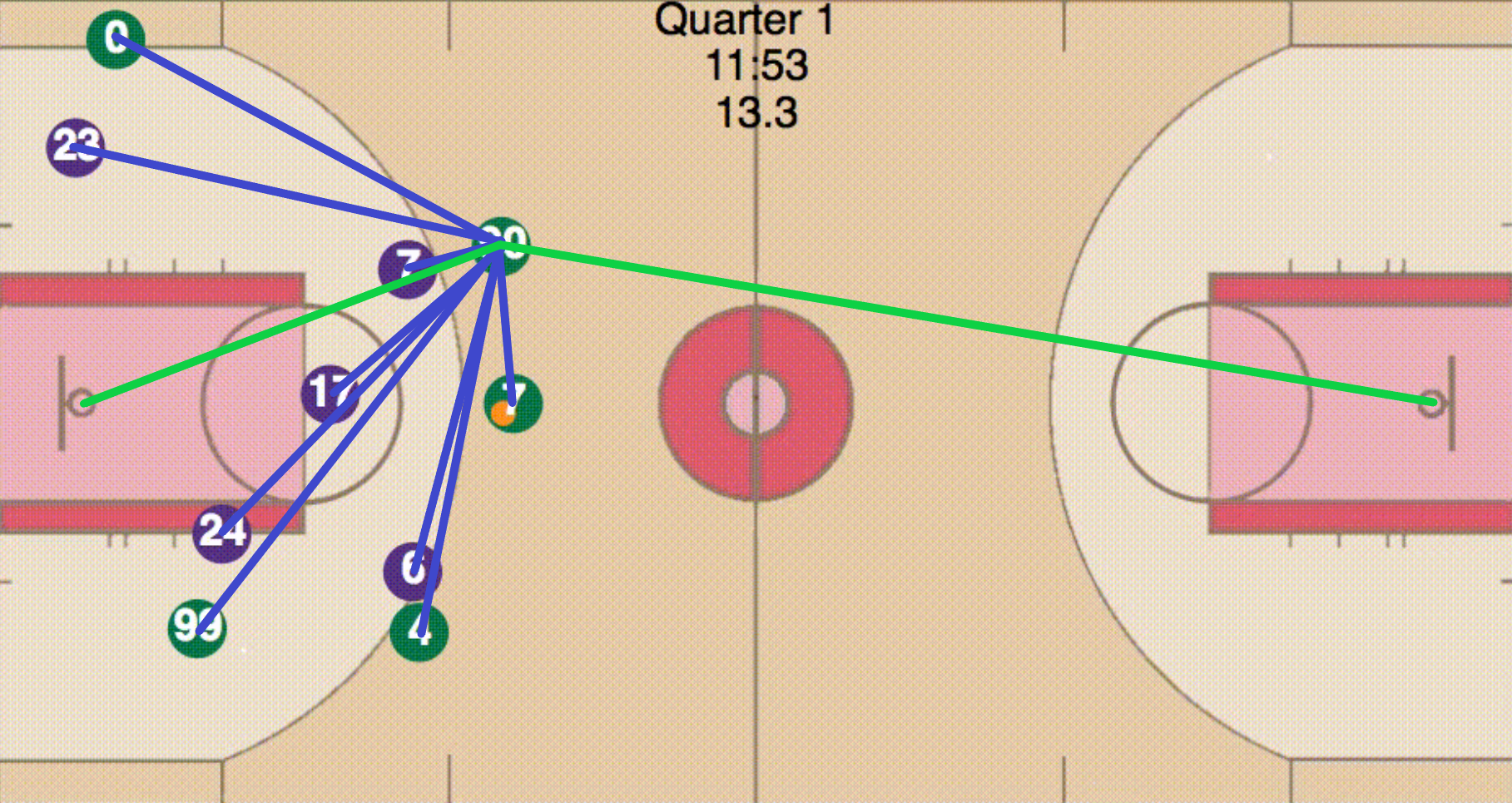}
    \caption{The game court in the NBA dataset with context.}
    \label{fig:contextInfo}
\end{figure}

We implemented a {two-layer LSTM} model, where the successive arrangement of two LSTM layers enhances the model's ability to capture multi-layered dependencies. The first layer extracts basic patterns from the input data and passes them to the subsequent layer, which models deeper and more complex dependencies. This layered structure enables the model to capture both short-term and long-term dependencies, thereby improving the accuracy of trajectory predictions. Model performance is further enhanced by providing supplementary information about the environment, see Fig.~\ref{fig:contextInfo}.

The model utilizes data based solely on the \textit{velocities} of the target object, teammates, and opponents. However, to provide a more comprehensive understanding of the environment, additional context information regarding the \textit{distances} to neighboring objects is incorporated. The environment comprises both dynamic elements, such as players, and static fixed points. In the case of the NBA dataset, these fixed points include the two basketball hoops, which serve as landmarks. A challenge in incorporating distances is that they can be arbitrarily large between objects. To address this, it is necessary to weight relevant distances more heavily to better represent valuable information. For this purpose, we apply a {transformation function} to scale the distances to a suitable value range:
\begin{equation}
    \label{distance_function}
    f(x) = sgn(x) \cdot 2 \cdot e^{-\frac{1}{2} \cdot |x|}.
\end{equation}
The transformation is applied to the distances to scale them into an appropriate value range. Its purpose is to optimize the influence of the distances on model predictions and enhance the informative value of the environmental properties. By prioritizing shorter distances, this transformation ensures that spatial relationships are more meaningful for the model. The resulting data is then organized into input-output pairs suitable for training the LSTM unit.

\begin{figure}[!bp]
    \centering
    \includegraphics[width=1\linewidth]{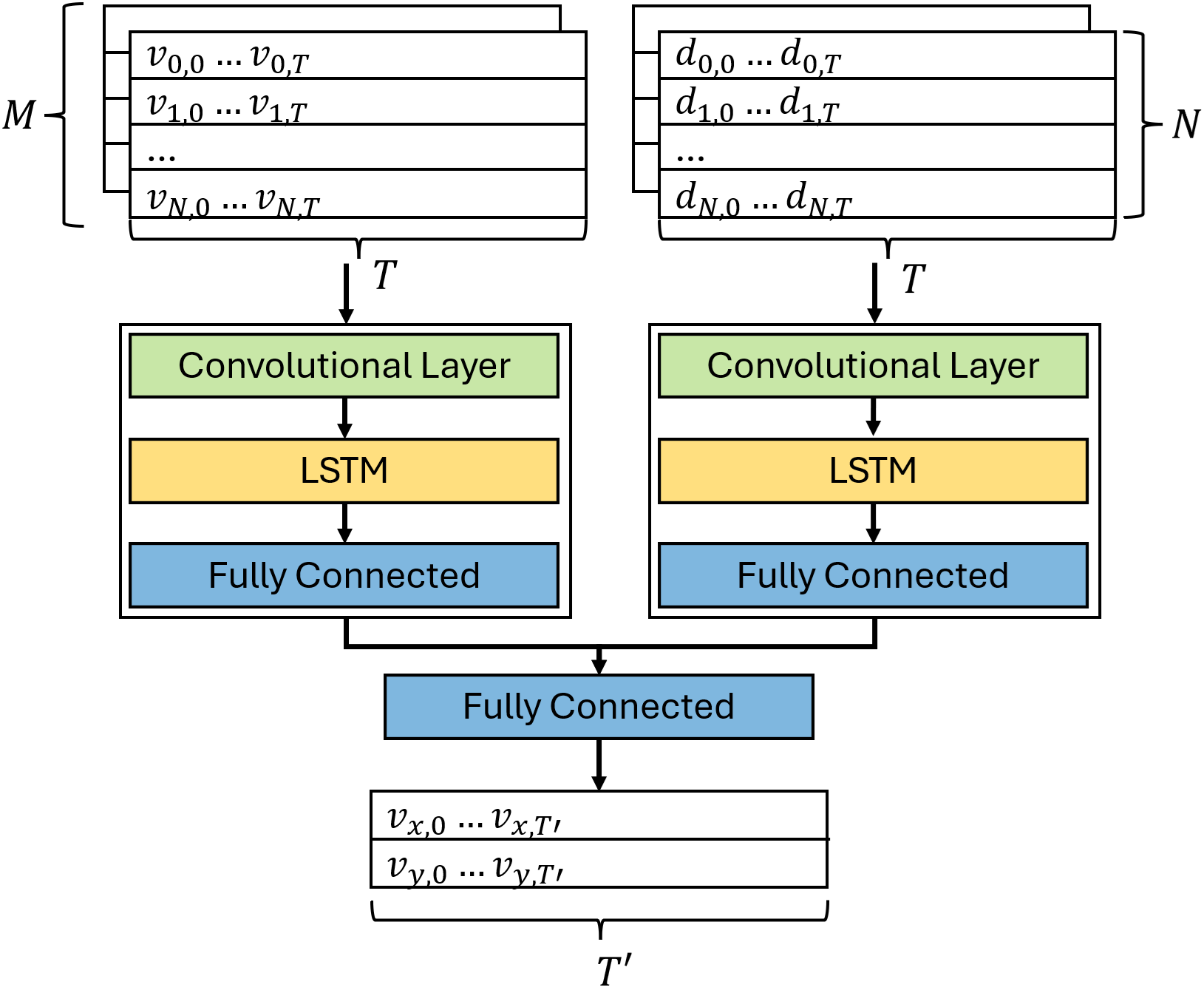}
    \caption{Pipeline of the CNN-LSTM model.}
    \label{fig:lstmCnnModel}
\end{figure}

\vspace{0.1cm}
\textbf{CNN-LSTM.} To efficiently prepare the context information for the LSTM-based model, a CNN is incorporated into the architecture (CNN-LSTM), as depicted in Fig.~\ref{fig:lstmCnnModel}. The convolutional layer is employed to extract relevant features from the input data, particularly from complex spatial contexts essential for trajectory prediction. By utilizing convolutional layers, the model can identify local dependencies and patterns in the data before these are processed by the temporal-sequential LSTM model. Here, $M$ represents the number of dimensions, $N$ denotes the number of objects (11 players, including 10 players and one ball), $T$ is the length of the input sequence, and $T^{'}$ is the length of the output sequence. $v$ corresponds to the {velocities} of the objects, and $d$ represents the distances to neighboring objects~\ref{fig:contextInfo}. The convolutional layer has a kernel size of ($2 \times 5$), operating along the time axis, processing two input channels, and producing eight output channels. A LeakyReLU activation function, with a negative gradient factor of 0.1, is applied between the convolutional layer and the subsequent LSTM. The LSTM consists of two layers (hidden size = 128), followed by a linear layer with 128 inputs and 100 outputs for encoding purposes. To integrate the context information, an additional linear layer is added, providing 400 inputs and 100 outputs for predicting the x- and y-coordinates over a two-second period. Note that, the described block represents a module list comprising four distinct information fields. The first field contains the \textit{velocity} information, the second field holds the transformed \textit{distances} to the nearest object, while the third and fourth fields represent the \textit{distances} to the two basketball hoops.

\vspace{0.1cm}
\textbf{LMU.}\label{sec:lmu} Although recurrent gating architectures, such as vanilla RNN, LSTM, and GRU, have proven empirically successful in managing hidden states, they do not always provide a direct mathematical foundation for how memory is formed and sustained over time. In contrast, LMUs adopt a more principled state-space approach, grounded in continuous-time filtering. Rather than relying primarily on learned gating mechanisms, LMUs utilize Legendre polynomials to effectively retain and transform past information, offering a theoretically sound framework to address the vanishing gradient problem and capture long-term dependencies.

The LMU~\cite{lmu} relies on approximating a continuous-time delay $F(s) = e^{-\theta s}$ using an ordinary differential equation (ODE) with window length $\theta$:
\begin{equation}
\theta \dot{m}(t) = Am(t) + B\,u(t),
\label{eq:ode_approx_short}
\end{equation}
where $m \in \mathbb{R}^d$ is the memory state, and $u \in \mathbb{R}$ is the input. The matrices $A \in \mathbb{R}^{d \times d}$ and $B \in \mathbb{R}^d$ encode Legendre polynomial coefficients. These matrices are calculated as
\begin{align}
    A &= \left[a_{ij}\right], \; a_{ij} = \begin{cases} 
    -1 & \text{if } i < j, \\
    (-1)^{i-j+1} & \text{if } i \geq j,
    \end{cases} \\
    B &= \left[b_i\right], \; b_i = (2i + 1)(-1)^i, \quad i, j \in [0, d-1].
\end{align}
The Equation~\eqref{eq:ode_approx_short} is discretized to obtain
\begin{align}
    m_t &= \Bar{A}m_{t-1} + \Bar{B}u_t,
\end{align}
where $\Bar{A}$ and $\Bar{B}$ are the discretized versions of $A$ and $B$, computed using methods such as {Euler's method} or {zero-order hold}. In a standard LMU layer, the memory state $m_t$ is updated using the discretized ODE, and a hidden state $h_t$ is computed by transforming both the input $x_t$ and the memory state by
\begin{align}
u_t &= e_x^T x_t + e_h^T h_{t-1} + e_m^T m_{t-1}, \\
m_t &= \Bar{A}m_{t-1} + \Bar{B}u_t, \\
h_t &= f\bigl(W_x x_t + W_h h_{t-1} + W_m m_t\bigr),
\end{align}
where $e_x$, $e_h$, and $e_m$ are encoding vectors and $W_x$, $W_h$, and $W_m$ are learnable weight matrices.

Our model utilizes two LMU layers. The first LMU layer transforms the input, which consists of 22 dimensions (11 objects, each with two features, i.e., \textit{positions} and \textit{velocities}), into a hidden size of $d$. A dropout layer is applied after the first LMU layer, followed by a second LMU layer with the same dimensionality. The final fully connected layer projects the output back to two dimensions, predicting the x- and y-coordinates over the forecast horizon ($T'$). The hyperparameter $d$ represents the dimensionality of the memory state $m$ and the size of the state-space matrices $A$ and $B$. In this model, $d$ is set to 256, providing sufficient capacity to capture long-term dependencies while ensuring computational efficiency. The parameter $\theta$ defines the window length for the continuous-time delay approximation, controlling how far into the past the memory state $m$ retains information. For this model, $\theta$ is set to 25, balancing the ability to capture temporal dependencies with model complexity. This configuration ensures that the LMU effectively represents sequential patterns in the input trajectory while maintaining efficiency.

\begin{figure}[!tp]
    \centering
    \includegraphics[width=1\linewidth]{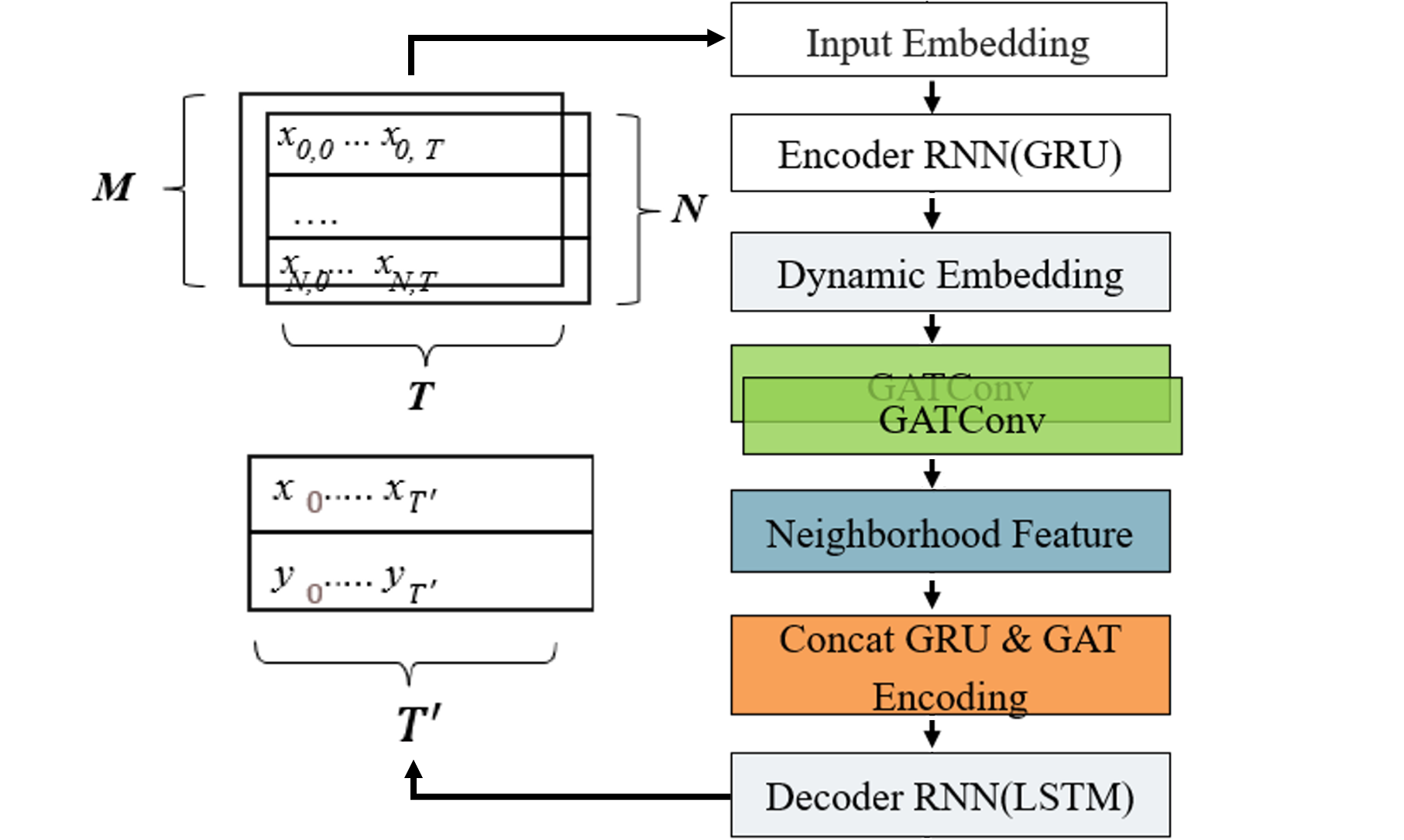}
    \caption{Pipeline of the GNN model.}
    \label{fig:fusion_pipeline}
\end{figure}

\vspace{+0.1cm}
\textbf{GNN.} To capture temporal dependencies, we enhance our GNN by integrating Graph Attention (GATs)~\cite{mo2021graph,GAT}. This integration enables the rich temporal representations generated by a RNN-based model to be processed dynamically in a graph-structured format, allowing GATs to emphasize the most relevant temporal and spatial interactions. GATs, a class of GNNs, utilize attention mechanisms to learn from graph-structured data. By dynamically assigning varying importance weights to neighboring nodes, GATs enable the model to focus on the most pertinent information for each node in the graph. The attention mechanism in GATs is computed using the following attention and aggregation functions: For the {node-level attention}, a given node $i$, the attention coefficients $\alpha_{i,j}$ between node $i$, and its neighbor $j$ are calculated as
\begin{equation}
    \resizebox{0.88\linewidth}{!}{$
    \displaystyle
    \alpha_{ij} = \frac{\exp\left(\text{LeakyReLU}\left(a^\top [W h_i \oplus W h_j]\right)\right)}{\sum_{k \in \mathcal{N}(i)} \exp\left(\text{LeakyReLU}\left(a^\top [W h_i \oplus W h_k]\right)\right)},\}
    $}
\end{equation}
where $h_i$ and $h_j$ are the feature vectors of nodes $i$ and $j$, $W$ is a learnable weight matrix, $a$ is the attention weight vector, $\oplus$ denotes concatenation, and $\mathcal{N}(i)$ is the set of neighbors of node $i$. {Weighted aggregation}: we compute the updated feature for node $i$ as a weighted sum of its neighbors' features:
\begin{align}
    h'_i = \sigma \Big(\sum_{j \in \mathcal{N}(i)} \alpha_{ij} W h_j \Big),
\end{align}
where $\sigma$ is an activation function (e.g., ReLU). For the {multi-head attention}, to enhance the model's capacity, GATs often employ multiple attention heads by computing
\begin{align}
    h'_i = \oplus_{k=1}^K \sigma\Big(\sum_{j \in \mathcal{N}(i)} \alpha_{ij}^{(k)} W^{(k)} h_j\Big),
\end{align} 
where $K$ is the number of attention heads.

GATs have demonstrated exceptional performance in some graph-based applications~\cite{Gatapplication}, providing a powerful mechanism to capture the intricate interactions between entities in dynamic environments. The flexibility of the attention mechanism allows GATs to learn more nuanced relationships compared to traditional GNNs. Our novel RNN-GAT-RNN-based encoder-decoder framework effectively addresses the challenges of predicting player movements in team sports. It combines RNNs and GATs to model both temporal dynamics and inter-player interactions, ensuring accurate predictions. 

Fig.~\ref{fig:fusion_pipeline} shows the overall architecture of the fusion model. The framework is organized into two primary stages: the {encoder} and the {decoder}. The {encoder} comprises two key components: (1) The {history encoder}, which utilizes a GRU-based RNN~\footnote{\label{refnote}{In a preliminary study, we optimized the architecture by exploring various RNN encoder-decoder combinations, including vanilla RNN, GRU, LMU, and LSTM. In this paper, we report only the best-performing configuration.}} to capture sequential dependencies within the historical trajectories of players. Each player’s trajectory, represented by x- and y-coordinates over time, is processed through the GRU to generate a latent representation that encodes both individual movement patterns and temporal dynamics. This provides the model with a comprehensive understanding of past movements, including position changes, velocity variations, and directional shifts. (2) The {interaction encoder} constructs an interaction graph where players are represented as nodes, and edges define relationships or interactions, such as distances or passes. A GAT processes this graph by calculating attention coefficients that determine the significance of neighboring players' movements for each target player. This enhances the representation of inter-player dynamics, producing interaction features that reflect each player’s role and influence within the team. The {decoder} stage consists of a decoder that employs an LSTM-based RNN\footref{refnote} to predict future trajectories based on the encoded features from the previous stage. Inputs to the decoder include dynamic features from the history encoder and interaction features from the GAT. These inputs are concatenated into a single vector, providing a comprehensive representation of the target player’s dynamics and interactions. The decoder then iteratively predicts the target player’s next position at each timestep, ultimately generating the entire trajectory. The processing begins with the data representation
\begin{align}
    \text{data} = \{ H_t, E_t, y_t \},
\end{align}
where $H_t$ represents the historical tracks of all players, $E_t$ is the edge set defining the graph structure, and $y_t$ is the ground truth trajectory of the target player. The interaction graph at time $t$ is processed as
\begin{align}
    G_t = \text{GNN}_{\text{inter}}(R_t, E_t),
\end{align}
where $R_t$ represents the sequential features \textit{position} and \textit{velocity} of all players at time $t$, and $E_t$ defines the graph structure at that timestep. To predict future trajectories we combine dynamics $r_{t_0}$ and interaction features $g_{t_0}$ as
\begin{align}
    f_{t_0} = \text{RNN}_{\text{fut}}([g_{t_0}, r_{t_0}]),
\end{align}
$f_{t_0}$ represents the predicted trajectory of a target player.
{Fusion} and {residual layers} further improve predictive accuracy by integrating feature fusion and residual connections. Temporal and relational features are concatenated, merging historical and interaction data, while residual connections reintroduce the original input features at later stages. This architecture allows the model to capture complex patterns effectively and mitigates vanishing gradients.

\begin{figure}[!t]
    \centering\includegraphics[width=0.9\linewidth]{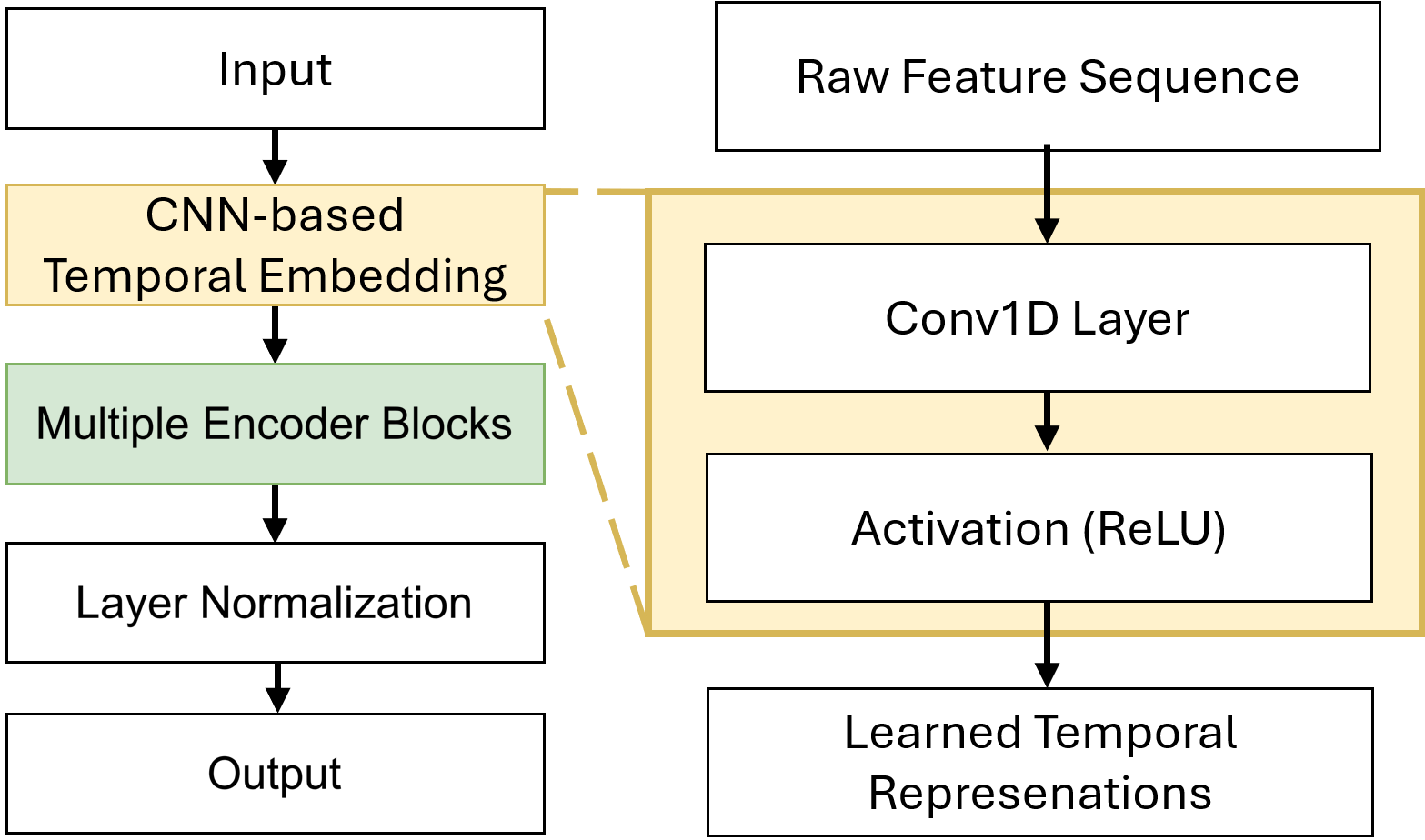}
    \caption{Architecture (left) with positional encoding (right) of our Transformer model; information flows from top to bottom.}
    \label{fig:transformer_architecture}
\end{figure}

\vspace{+0.1cm}
\textbf{Transformer Model.} RNNs and their variants have demonstrated efficacy in modeling sequential data, owing to their ability to capture temporal dependencies. However, their inherent sequential processing limits computational efficiency and impedes the modeling of long-range dependencies in extended sequences. Although GNNs, especially GATs, excel in modeling spatial relationships by dynamically attending to relevant graph nodes, they seem less effective in capturing temporal information over prolonged time periods.

Fig.~\ref{fig:transformer_architecture} shows the architecture of the transformer model. The input sequence is processed through an encoder-only transformer, which mitigates the error accumulation commonly encountered in decoder-based transformer models. This encoder simultaneously processes the entire sequence, utilizing positional embeddings to represent the temporal order of the input data. 
Specifically, a CNN-based temporal embedding module is used to extract local sequential patterns from the input (\textit{velocity} vectors and \textit{positions}). A 1D convolutional layer with a fixed kernel size transforms the raw features into a richer representation space, captures implicit temporal dependencies and preserves the order. This method increases the feature capacity available to the attention mechanism and stabilizes the learning process. Velocities effectively capture motion trends and relative dynamics. However, we found that absolute positions provide essential spatial context that enables the model to better understand formations, spacing, and positional roles. We think, this input strategy is effective for transformer architectures due to their global receptive field. Instead, most recurrent models such as vanilla RNN, GRU, and LSTM perform best on relative features, e.g., velocities.

\begin{figure}[!t]
    \centering
\includegraphics[width=0.9\linewidth]{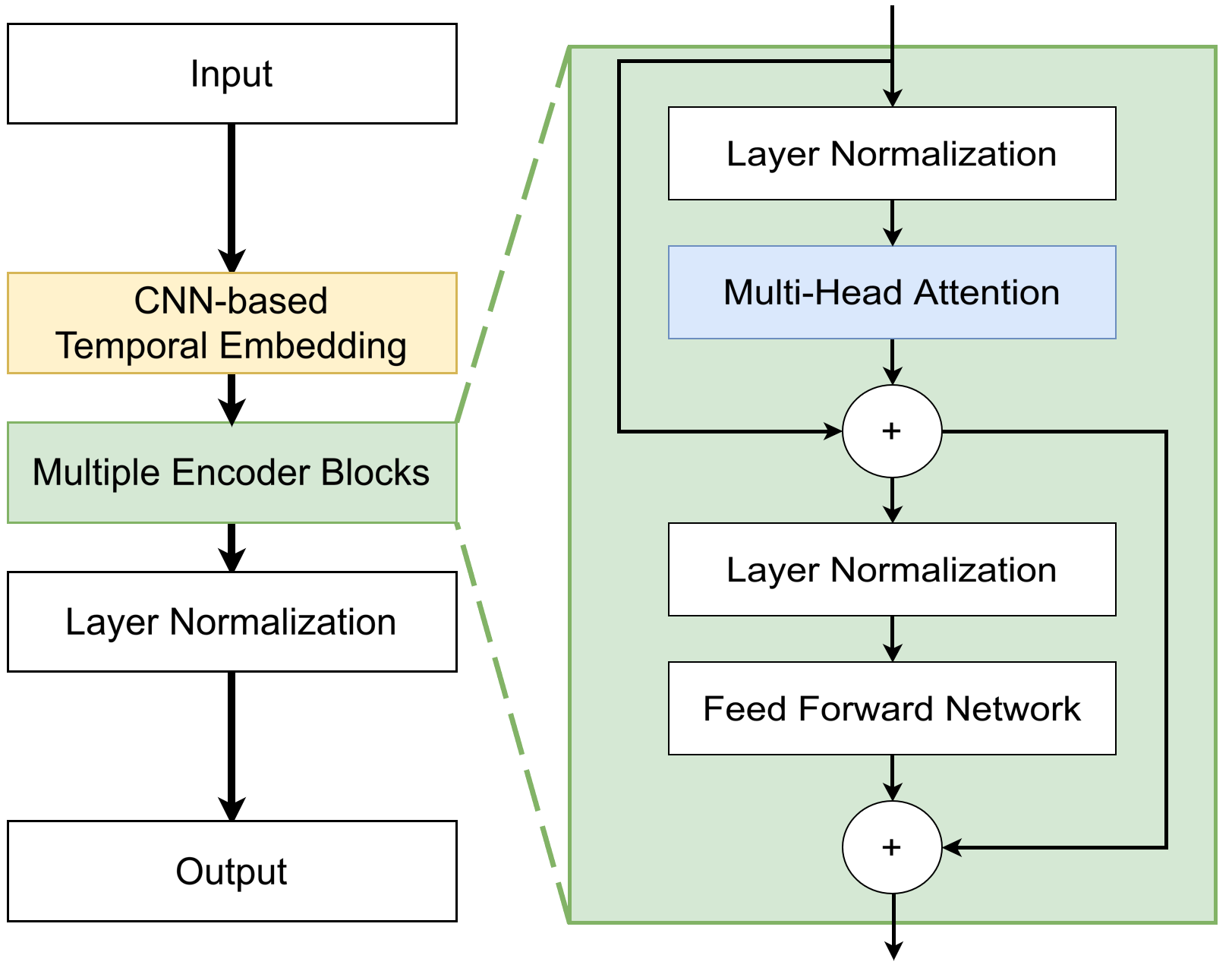}
    \caption{Visualization of the data flow of an encoder block.}
    \label{fig:encoder_block}
\end{figure}

Each encoder block, see Fig.~\ref{fig:encoder_block}, comprises three main components: a multi-head attention (MHA) mechanism, a feed-forward network (FFN), and layer normalization. The MHA mechanism enables the model to simultaneously attend to different parts of the input sequence, effectively capturing both short-range and long-range dependencies. This is followed by a position-wise FFN with a hidden size of 1,024, which processes the output at each position independently. Residual connections are applied after both the MHA and FFN layers to stabilize the training process and enhance gradient flow. Layer normalization is employed to ensure stability and normalize the data as it propagates through the network.

The MHA mechanism utilizes eight attention heads ($\text{num}. \text{heads} = 8$) and is computed by
\begin{align}
    \text{Attention}(Q, K, V) &= \text{softmax}\left(\frac{QK^\top}{\sqrt{d_k}}\right)V,
\end{align}
where $Q$, $K$, and $V$ are the query, key, and value matrices, respectively. The dimension of each key vector is $d_k = 4$. These matrices are derived from the input sequence $X$ through linear transformations
\begin{align}
    Q = XW_Q, \quad K = XW_K, \quad V = XW_V,
\end{align}
with $W_Q$, $W_K$, and $W_V$ as learnable weight matrices. We pass the output of the MHA to the feed-forward network
\begin{align}
    \text{FFN}(x) = \sigma(xW_1 + b_1) W_2 + b_2,
\end{align}
where $W_1$, $W_2$, $b_1$, and $b_2$ are learnable parameters, and $\sigma$ is an activation function, typically ReLU.

The transformer model consists of six encoder blocks (\text{num}. \text{encoder} \text{blocks} $ = 6$). Each block has an input embedding size of $d_\text{model} = 256$, which matches the dimensionality of the input embeddings and positional encodings. The input sequence length is determined by the trajectory data, with each sequence processed entirely by the encoder.

To train the model effectively, the {AdamW optimizer} is utilized, incorporating weight decay to mitigate overfitting. The learning rate follows a warm-up schedule:
\begin{align}
    \text{lr}(t) = \frac{d_\text{model}^{-0.5} \cdot t}{\text{warmup steps}^{1.5}},
\end{align}
where $d_\text{model} = 256$ and $t$ is the current step. The number of \text{warmup} \text{steps} can be selected according to the specific training regimen. By utilizing these components, including an eight-head MHA mechanism, an FFN with a hidden size of 1,024, six encoder blocks, and a robust training strategy, the transformer model is highly effective in capturing complex player-ball interactions and long-range temporal dependencies.

\subsection{Postprocessing}\label{postprocessing}

For models that output relative directed velocities $v$ instead of absolute positions $p$, we reconstruct the position sequence $p_{i,n}$ from the predicted velocities $v$. This reconstruction relies on a fixed time step of $ \Delta t=0.04s$ between consecutive positions $p_{i-1}$ and $p_{i}$. Each forecasting process begins from a known initial position $p_{i-1}$, which serves as the starting point for iteratively computing future positions: $p_{i} = p_{i-1} + v_{i} \Delta t$.

\section{Experiments}
\label{sec:experiments}

Sec.~\ref{section_exp_dataset} provides a detailed description of the dataset, while Sec.~\ref{section_exp_setup} outlines the experimental setup. Error metrics are summarized in Sec.~\ref{section_exp_metrics}. The benchmark experiments were conducted on a high-performance system equipped with Tesla V100-SXM2 GPUs, and the models were implemented using \textit{PyTorch} (version 2.4.1).

\subsection{Dataset}
\label{section_exp_dataset}

We utilize the publicly available NBA dataset~\cite{NBADataset}, which includes data from the 2015-2016 season, comprising 508 hours of playtime recorded at 25\unit{fps}. To address potential recording errors, interpolation was applied for data smoothing. The dataset is partitioned into three subsets: 70\% for training, 20\% for validation, and 10\% for testing. This partitioning ensures that no games are shared across subsets, enabling the model to be evaluated on completely unseen data for an unbiased performance assessment.

\subsection{Experimental Setup}
\label{section_exp_setup}

\textbf{Experiment 1: Optimal Input Length.} The first experiment seeks to determine the optimal input length for the models under evaluation. Specifically, it examines how varying the number of past timesteps used as input influences predictive accuracy by assessing changes in prediction error for different input sequence lengths. {The dataset comprises 128,538 trajectories, each containing 50 predicted timesteps}, with various input lengths tested to identify the ideal historical context for accurate forecasting. Multiple input sequence lengths are evaluated to determine the point at which increasing the input length improves accuracy up to a certain threshold, beyond which further inclusion of timesteps either yields only marginal improvements or potentially degrades performance due to overfitting or increased uncertainty. {In a preliminary study, we evaluated the effects of various input horizons (up to 10\unit{s}), i.e., \textbf{0 to {2}} in steps of 0.05\unit{s}, and \textbf{4}, 6, 8, and 10\unit{s}, for each model individually. Note that for computationally inefficient models such as GNN and Transformer, we only evaluated the bolded horizons, as no performance improvement was observed beyond 2\unit{s}.} This strategy offers insights into the minimal historical context needed for accurate predictions while minimizing computational overhead. The results from this experiment inform the selection of input length for subsequent evaluations of other models.

\textbf{Experiment 2: Generalization Within a Unique Team.} The second experiment assesses the models' ability to generalize to unseen games while remaining within the same team context. {Again, we employ the same dataset with 128,538 trajectories, each containing 2\unit{s} input- and forecast data.}\footnote{\label{refnote2}{We found that no model benefits from input lengths greater than 2\unit{s}, and the prediction error increases significantly beyond a 2\unit{s} forecast horizon.}} The models are trained on a subset of games from a single team, with the test set comprising games from the same team that were excluded from the training process. This setup evaluates how well the models can predict player trajectories in new game scenarios, having learned the team's specific playing style and dynamics during training. By fixing the input length to 50 timesteps, the experiment establishes consistent conditions for assessing the models' ability to generalize to new yet familiar contexts, thereby emphasizing their robustness in capturing team-specific movement patterns.

\textbf{Experiment 3: Cross-Team Generalization.} This experiment evaluates the models' ability to generalize across different teams by training on data from one team and testing on entirely unseen games from another team. The dataset employs 48,274 trajectories, each with a fixed input length of 50 timesteps and an output length of 50 timesteps, i.e., 2\unit{s}.\footref{refnote2} This experiment exposes the models to unfamiliar team dynamics, simulating real-world scenarios where the system must adapt to diverse strategic and competitive environments. The results propose applicability of the learned motion patterns and the models' robustness in handling varied team behaviors.

\subsection{Error Metrics}
\label{section_exp_metrics}

Performance is evaluated using both spatial and angular metrics. Specifically, the final displacement error (FDE) and the average displacement error (ADE) are measured in meters, while the final angular error (FAE) and the average angular error (AAE) are measured in degrees. These metrics collectively assess how closely the predicted trajectory approximates the ground truth in terms of both position and orientation. The FDE and ADE metrics are defined as follows:
\begin{equation}
    \text{FDE} = \frac{1}{N} \sum_{i=1}^{N} \| y_{i,T} - \widehat{y}_{i,T} \|,
\label{eq:FDE}
\end{equation}
\begin{equation}
    \text{ADE} = \frac{1}{N\cdot T} \sum_{i=1}^{N}\sum_{t=1}^{T}\left\| y_{i,t} - \widehat{y}_{i,t} \right\|,
\label{eq:ADE}
\end{equation}
where $\widehat{y}_{i,t}$ denotes the predicted position at timestep $t$ for sample $i$, $y_{i,t}$ is the corresponding ground truth, $T$ is the forecasting horizon, and $N$ is the total number of samples. Here, ADE quantifies the average positional mismatch at every timestep, whereas FDE focuses on the final position at $t=T$.

For angular errors, orientation is derived from velocity vectors of the ground truth trajectory $\vec{v}_y$ and the predicted trajectory $\vec{v}_{\widehat{y}}$. First, the cosine of the angle between these vectors is computed by
\begin{equation}
    \Bar{\theta} = \arccos \Big( \frac{\vec{v}_y \cdot \vec{v}_{\widehat{y}}}{\|\vec{v}_y\|\cdot \|\vec{v}_{\widehat{y}}\|} \Big).
\end{equation}
A signed angle $\theta$ is then obtained to distinguish leftward and rightward orientations: 
\begin{equation}
    \theta = 
    \begin{cases}
        \Bar{\theta} & \text{if} \,\,\,\,\, \vec{v}_y \times \vec{v}_{\widehat{y}} > 0,\\
        -\Bar{\theta} & \text{else}.
    \end{cases}
\end{equation}
To measure orientation accuracy over time, we define AAE as
\begin{equation}
    \text{AAE} = \frac{1}{N \cdot T} \sum_{i=1}^N \sum_{t=1}^T \bigl|\theta_{i,t}\bigr|,
\end{equation}
and the FAE is defined as
\begin{equation}
    \text{FAE} = \frac{1}{N} \sum_{i=1}^N \bigl|\theta_{i,T}\bigr|,
\end{equation}
where $\theta_{i,t}$ is the signed angle between the ground truth and predicted velocity vectors for sample $i$ at time $t$. Accordingly, AAE captures the average angular discrepancy at each timestep, and FAE captures the final angular mismatch at $t=T$. These spatial and angular metrics furnish a comprehensive assessment of trajectory prediction performance.
\section{Results}
\label{sec:results}

We evaluated all models based on the impact of input history length on the forecasting horizon (Sec.~\ref{sec:exp:1}), generalization capability, forecasting uncertainty, computational complexity, and the ability to leverage implicit contextual information derived from team dynamics (Sec.~\ref{sec:exp:2} - \ref{sec:exp:3}). For all experiments, we evaluated the following models: two classic baselines, a Constant Velocity model and a Linear model, as well as our TCNN, LSTM, CNN-LSTM, LMU, GNN, and Transformer architectures.

\begin{figure*}[!bthp]
    \centering
    \includegraphics[trim={0 0 0 50},clip,width=1\linewidth]{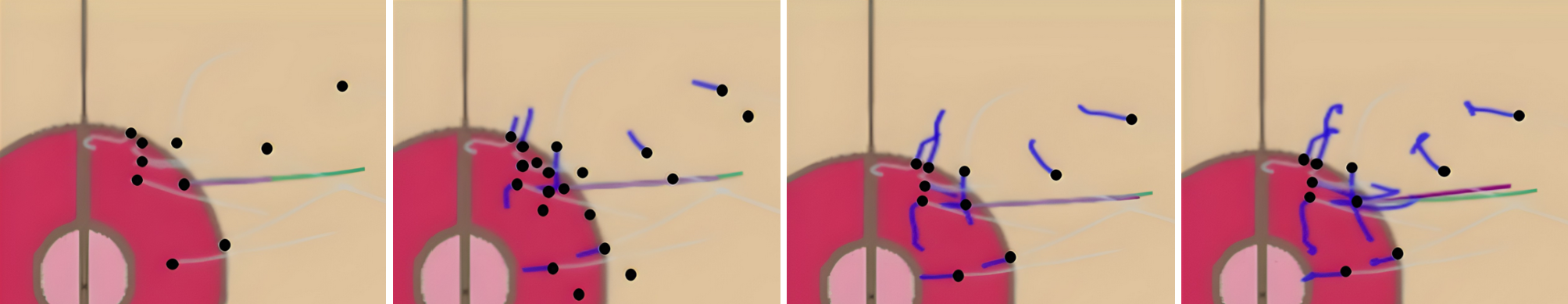}
    \caption{Comparison of trajectory forecasting for different input lengths (from left to right: 0.04\unit{s}, 1\unit{s}, 2\unit{s}, and 4\unit{s}). Best results are when the purple curve covers the green curve, e.g., at 2\unit{s} input length.}
    \label{fig:input_len_example}
\end{figure*}

\subsection{Optimal Input Features}\label{sec:exp:0}

For a fair comparison, we optimized each model individually w.r.t. input features (\textit{positions}, \textit{velocities}, \textit{distances}). We found that Constant Velocity, Linear, TCNN, LSTM, and CNN-LSTM perform best on \textit{velocities}\footref{fn_initialpos}. Instead, LMU, GNN, and Transformer exploited also \textit{positions}. 

\begin{figure}[!bhp]
    \centering
    \includegraphics[width=1\linewidth]{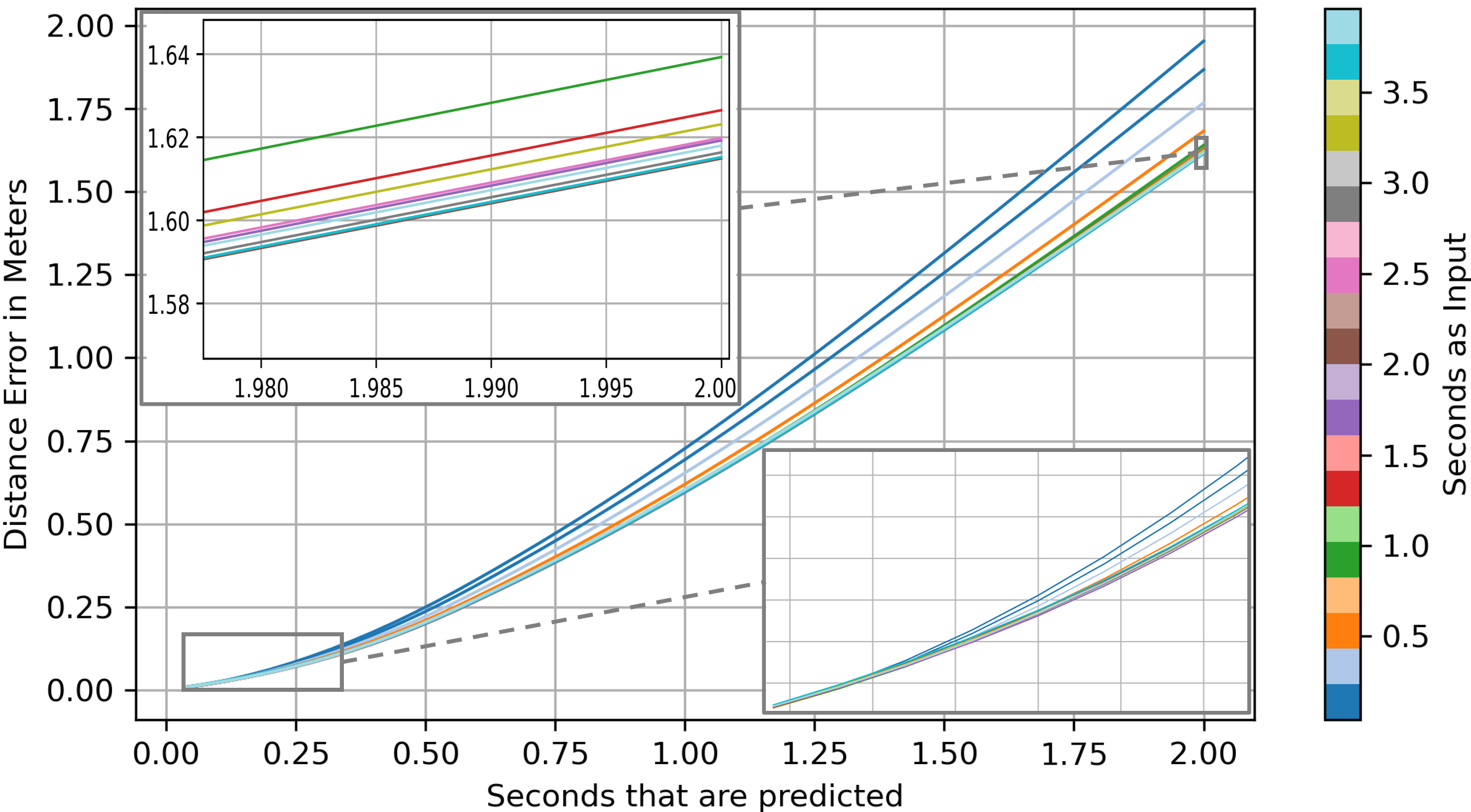}
    \caption{Evaluation error (in \textit{m}) of different input lengths.}
    \label{fig:input_len}
\end{figure}

\subsection{Experiment 1: Optimal Input Length}\label{sec:exp:1}

This experiment investigates the effect of varying the input sequence length on the accuracy of trajectory prediction. Fig.~\ref{fig:input_len} presents the results, where the horizontal axis represents the forecasting horizon (in seconds) and the left vertical axis denotes the distance error (in meters). Each colored line corresponds to a different input sequence length. The data reveals that shorter input windows, e.g., less than 2\unit{s}, result in significantly higher error rates, as the model lacks sufficient historical context to make accurate predictions (dark blue colored lines in the top right that represent an input length of 0.15\unit{s}). As the input length increases, the error decreases 1.62\unit{m} until it levels off at a plateau around 3.5\unit{s} (light blue lines in the top-left sub-graph in Fig.~\ref{fig:input_len}), indicating that a broader temporal context improves prediction accuracy. Beyond 2\unit{s}, however, the models no longer benefit significantly from additional input data; the error curve begins to level off, suggesting that further increases in the input window yield diminishing returns in terms of predictive accuracy. For simplicity, Fig.~\ref{fig:input_len} only shows the results of the LSTM model. We observed similar results across all other model architectures: increasing the input length up to 2\unit{s} consistently reduces error, while additional expansion results in minimal improvements. Thus, we suggest an optimal input length of around 2\unit{s} for most models, providing a balance between accuracy and computational efficiency, in basketball. 
{Interestingly, at forecast horizons of 2\unit{s} and beyond, an increase in input length offers little benefit, see the top-left sub-graph, e.g., error variations remain minimal between 1.61 and 1.63\unit{m} at input lengths between 1.0 and 3.0\unit{s}. We believe this is due to the highly dynamic nature of basketball: players frequently change direction, making motion patterns beyond 2\unit{s} less predictive. In contrast, preliminary tests on soccer and pedestrian data suggest that significantly longer input horizons are beneficial in those domains, as movements are generally more stable and less prone to sudden backward or lateral changes.}

Fig.~\ref{fig:input_len_example} illustrates the same scenario with varying input lengths (0.04\unit{s}, 1\unit{s}, 2\unit{s}, and 4\unit{s}). The green line represents the ground truth, while the purple line indicates the predicted trajectory. The black dots denote the positions of individual players, the blue path corresponds to the input trajectory, and the gray lines depict the projected positions of the players over the next two seconds. We observed that the error of the estimated trajectory increases when the input length is significantly smaller than 2\unit{s}.

\subsection{Experiment 2: Generalization Within a Unique Team}\label{sec:exp:2}

\begin{figure*}[!t]
    \centering
	\begin{minipage}[t]{0.493\linewidth}
        \centering
    	\includegraphics[width=1\linewidth]{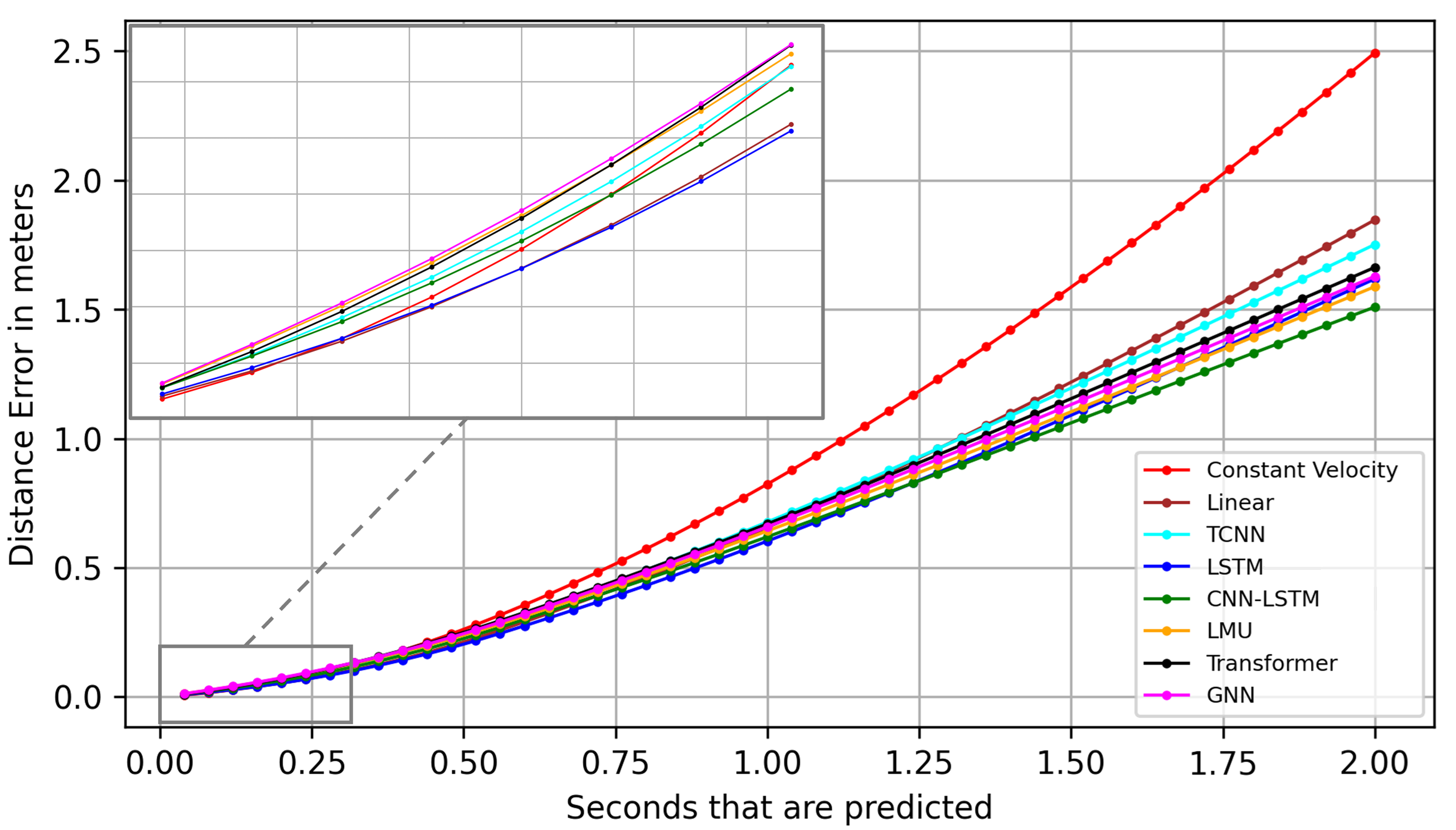}
        \subcaption{Trajectory forecasting (in \textit{m}).}
        \label{fig:distanceErrorLukas1}
    \end{minipage}
    \hfill
	\begin{minipage}[t]{0.493\linewidth}
        \centering
    	\includegraphics[width=1\linewidth]{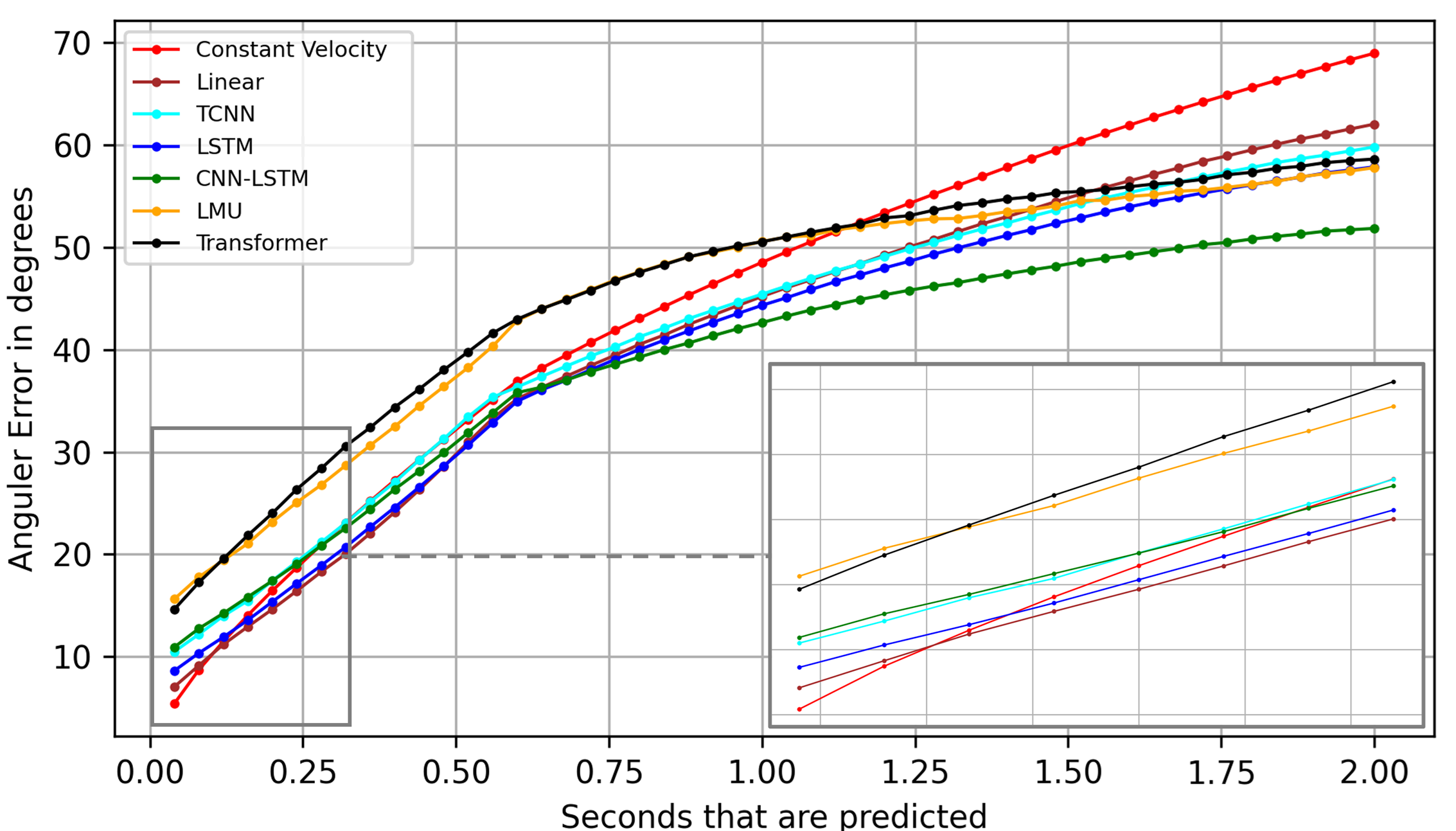}
        \subcaption{Orientation forecasting (in $^{\circ}$).\footref{foot_elbow}}
        \label{fig:distanceErrorLukas2}
    \end{minipage}
    \caption{Evaluation of all models. An optimized constant velocity KF works best below 0.12,\textit{s}, after which the dynamic motion becomes nonlinear and ML models are necessary.}
    \label{fig:distanceErrorLukas}
\end{figure*}

This experiment evaluates the performance of all methods w.r.t. trajectory forecasting on unknown input data, i.e., train and test on a specific team (Lakers). Fig.~\ref{fig:distanceErrorLukas} illustrates the distance error for each model architecture.\footnote{Note that the curves in Fig.~\ref{fig:distanceErrorLukas2} exhibit a distinct elbow. We believe this is likely attributable to characteristics inherent to the dataset or to specific patterns in NBA playstyle. When applying our methods to similar datasets from other domains, such as soccer or pedestrian trajectories, this effect did not occur. Notably, the Constant Velocity baseline also displays this pattern, further suggesting a domain-specific cause.\label{foot_elbow}}

For short forecast horizons below 0.5\unit{s}, all models yield significantly lower errors than for long ones.
The results show that even the Constant Velocity and Linear neural network models perform well when motion dynamics and trajectory changes are minimal, i.e., below 0.5\unit{s} forecast horizons, see Tbl.\ref{table:tableExp2_small} and the top-left sub-graph in Fig.\ref{fig:distanceErrorLukas1}.
Specifically, the Linear model achieved an FDE of 0.20\unit{m} and an FAE of 28.59$^{\circ}$.
The sub-graph indicates that up to a 0.12\unit{s} forecast horizon, both Constant Velocity and Linear models (brown and red curves) outperform all others.
However, between 0.12\unit{s} and 0.30\unit{s}, the Constant Velocity model fails to capture non-linear motion, and the Linear and LSTM models begin to outperform the rest.
Beyond 0.4\unit{s}, all models outperform the Constant Velocity model.
And beyond 1.3\unit{s}, all models outperform the Linear model.
We believe this is as the Constant Velocity and Linear models benefit the least from additional context information, as they are unable to capture temporal dependencies due to the lack of recurrent or attention mechanisms.

{Out of all methods, our CNN-LSTM benefits the most from supplementary context information. Its combination of CNN and LSTM outperforms pure LSTMs with increasing forecast horizon starting at 1.28\unit{s}, by up to 11\unit{cm} at 2\unit{s} forecast horizon, see Tbl.\ref{tableExp2}}. 
Its distance error (green curve) decreases by 40\% from 2.5\unit{m} (Constant Velocity) to 1.5\unit{m}, see Fig.~\ref{fig:distanceErrorLukas1}. 
And its orientation error reduces by 25\% from 68.98\,$^{\circ}$ to 51.86\,$^{\circ}$, see Fig.~\ref{fig:distanceErrorLukas2}. 
Instead, the LMU model achieves a slightly higher FDE of 1.59\unit{m}. {Thus, we think that Legendre polynomials may be worth investigating as part of a hybrid CNN-LMU architecture.}
Our GNN is almost on-par with LMU and slightly outperforms the Transformer model. 
We think that this is as GNN exploits both GAT and graph-context. So, it returns an FDE of 1.62\unit{m}. 
Interestingly, the Transformer model performs slightly worse (FDE = 1.66\unit{m}) than the recurrent (LSTM, LMU, CNN-LSTM) and GNN models. 
We believe that our Transformer model suffers from having too little data for effective training.

\begin{table}[!t]
\centering
\caption{Errors at a forecast horizon of 2.0\unit{s}.}
\label{tableExp2}
\begin{tabular}{l|cccc}
\textbf{Model} & \textbf{ADE [\textit{m}]} & \textbf{FDE [\textit{m}]} & \textbf{AAE [$^{\circ}$]} & \textbf{FAE [$^{\circ}$]} \\ \hline
Constant Velocity & 0.99 & 2.49 & 45.42 & 68.98 \\
Linear            & 0.76 & 1.85 & 41.76 & 62.05 \\
TCNN              & 0.76 & 1.75 & 42.23 & 59.83 \\
LSTM              & 0.68 & 1.62 & 40.61 & 57.87 \\
CNN-LSTM          & \textbf{0.67} & \textbf{1.51} & \textbf{38.92} & \textbf{51.86} \\
LMU               & 0.70 & 1.59 & 45.38 & 57.80 \\
Transformer       & 0.74 & 1.66 & 46.12 & 58.64 \\
GNN               & 0.71 & 1.62 & 45.18 & 54.20 \\ 
\end{tabular}
\end{table}

In essence, our findings highlight the importance of selecting models based on the forecast horizon.
For very short horizons (up to 0.12\unit{s}), classical approaches such as a Constant Velocity Kalman Filter may be sufficient.
However, for forecast horizons beyond 0.12\unit{s}, we recommend using recurrent, graph-based, or attention-based models.
For horizons exceeding 0.76\unit{s}, at least an LSTM model should be employed, and beyond 1.3\unit{s}, a CNN-LSTM architecture is advisable.

\begin{figure*}[!t]
    \centering
	\begin{minipage}[t]{0.493\linewidth}
        \centering
    	\includegraphics[width=1\linewidth]{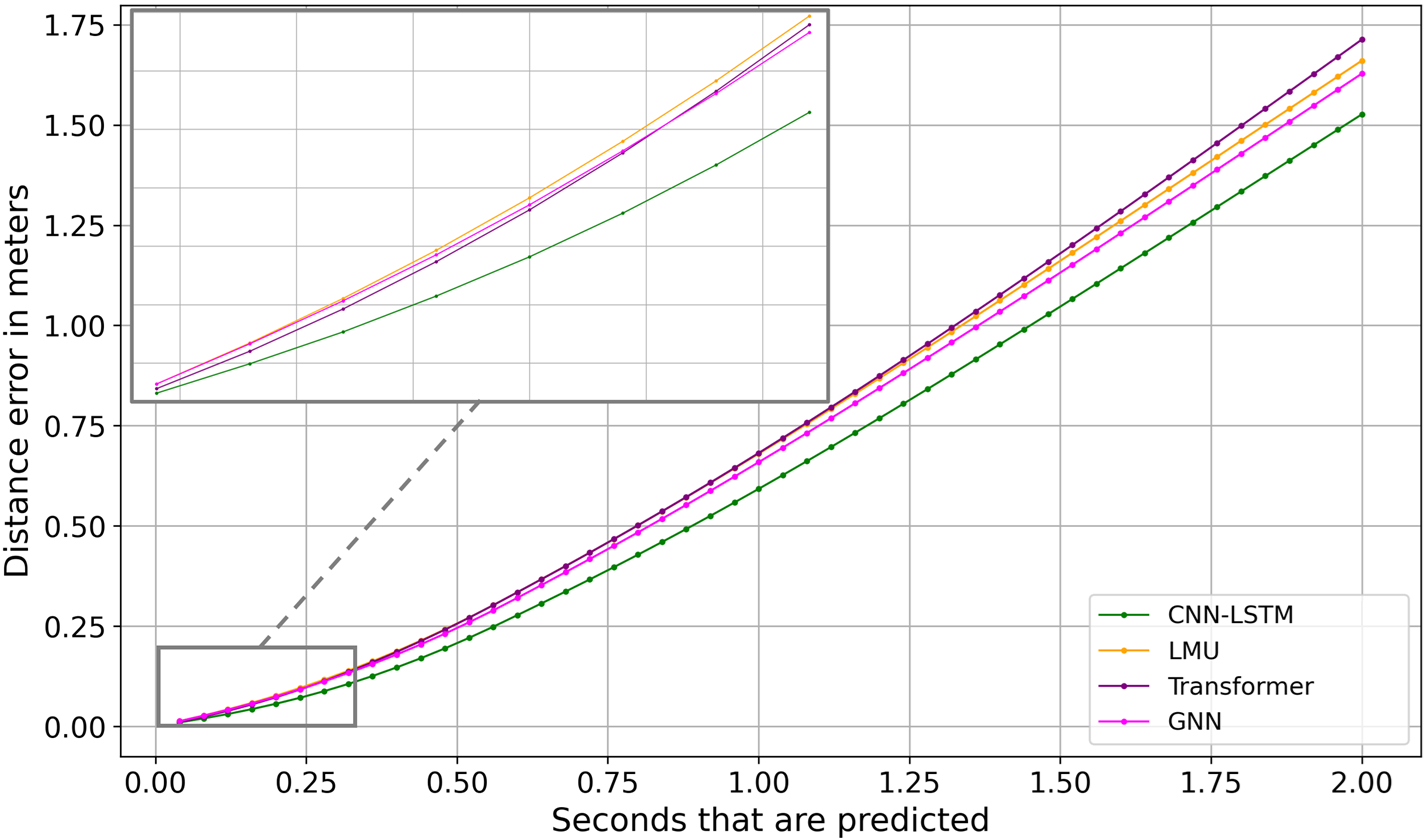}
        \subcaption{Trajectory forecasting (in \textit{m}).}
        \label{fig:genTestLukas1}
    \end{minipage}
    \hfill
	\begin{minipage}[t]{0.493\linewidth}
        \centering
    	\includegraphics[width=1\linewidth]{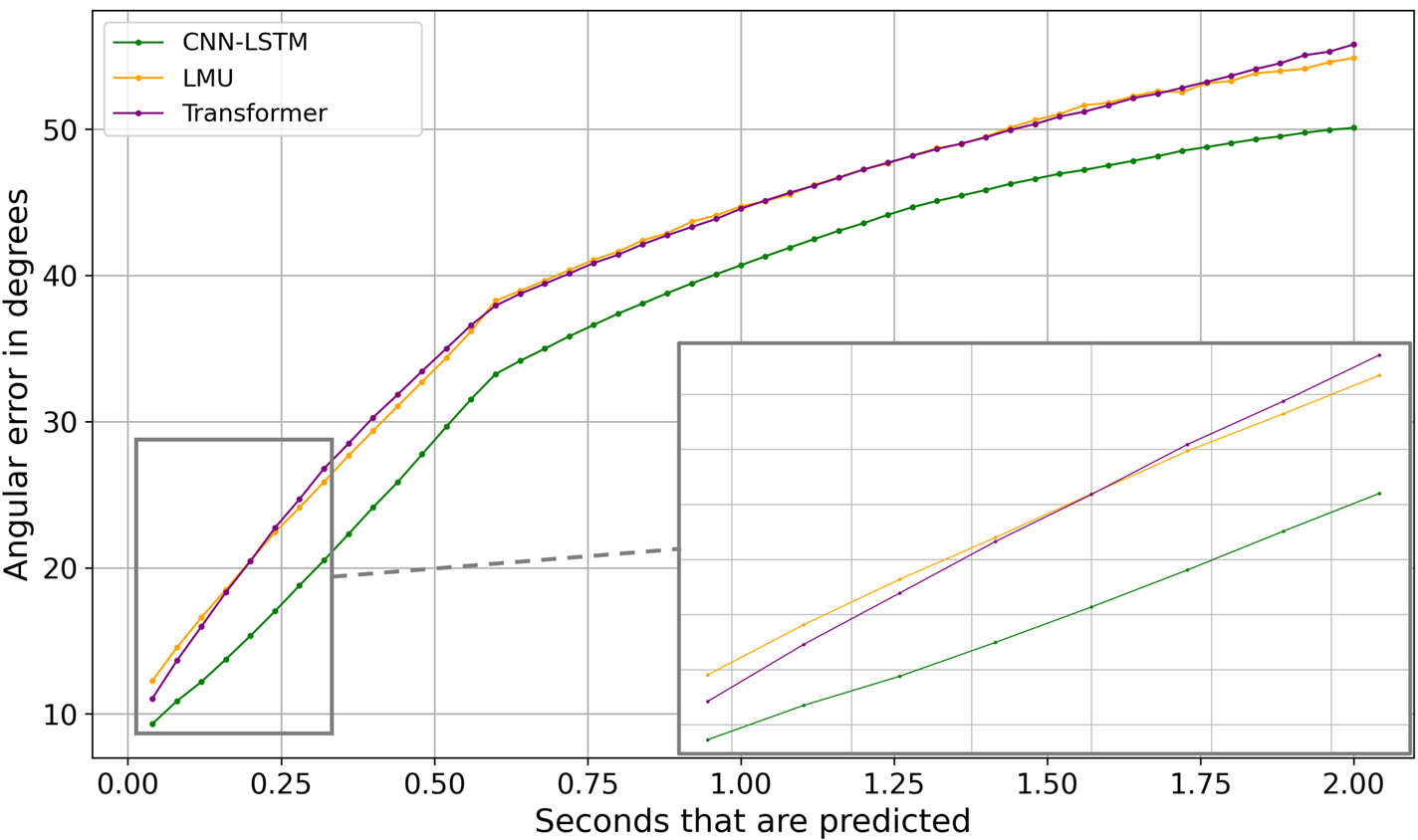}
        \subcaption{Orientation forecasting (in $^{\circ}$).}
        \label{fig:genTestLukas2}
    \end{minipage}
    \caption{Evaluation for unseen games.}
    \label{fig:genTestLukas}
\end{figure*}

\begin{table}[!t]
\centering
\caption{Errors at a forecast horizon of 0.48\unit{s}.}
\label{table:tableExp2_small}
\begin{tabular}{l|cccc}
\textbf{Model} & \textbf{ADE [\textit{m}]} & \textbf{FDE [\textit{m}]} & \textbf{AAE [$^{\circ}$]} & \textbf{FAE [$^{\circ}$]} \\ 
\hline
Constant Velocity & 0.10      & 0.24             & 19.31           & 31.22 \\
Linear       & \textbf{0.09}  & 0.20             & \textbf{17.56}  & \textbf{28.59}  \\
TCNN         & 0.10           & 0.23             & 20.47           & 31.28 \\
LSTM         & \textbf{0.09}  & \textbf{0.19}    & 18.25           & 28.60 \\
CNN-LSTM     & 0.10           & 0.21             & 20.20           & 29.94 \\
LMU          & 0.11           & 0.23             & 25.97           & 36.39 \\
Transformer  & 0.11           & 0.24             & 26.96           & 38.02 \\
GNN          & 0.11           & 0.23             & 22.10           & 32.85 \\
\end{tabular}
\end{table}

\subsection{Experiment 3: Cross-Team Generalization}\label{sec:exp:3}

In this experiment, we only benchmark the best-performing models (CNN-LSTM, LMU, Transformer, and GNN) to predict trajectories for new games featuring unknown players. Initially, the models were trained on data of a specific team. Although new trajectories were provided, the players within each team were always known. This setup enabled the model to learn patterns and correlations specific to a given team without being exposed to unknown player attributes.

Fig.~\ref{fig:genTestLukas} demonstrates the performance of our models trained on Lakers on an unknown match between Cleveland and Houston. The players in this match are also new to the model. The results show that the models are robust to unknown data, with errors nearly identical to those observed in earlier tests, compare Tbl.~\ref{resultExp3}. This suggests that the models generalize even when confronted with unfamiliar games and players. Tbl.~\ref{resultExp3} presents a comparison between the models of Experiment 2, that were trained and tested on a specific team (Lakers) to the results of Experiment 3. 

The minimal differences in error (CNN-LSTM: $\Delta$ FDE=0.02\unit{m}) between the two experiments indicate that the models do not overfit to specific players or teams. Instead, they demonstrate the ability to make robust, generalized predictions even on unknown (left out) data. For readability, this time we only plot methods that exploit context. Notably, our CNN-LSTM model (green curve) outperforms the more complex Transformer and GNN models across all forecast horizons. This suggests that a carefully engineered hybrid architecture may outperform attention-based models such as GNNs (with GAT) and Transformers, particularly in data-scarce settings and dynamic sports.

\begin{table}[!t]
\centering
 \caption{Difference in error between Experiment 2 and 3.}
\label{resultExp3}
\begin{tabular}{l|cccc}
\textbf{Model} & \multicolumn{1}{l}{\textbf{$\Delta$ADE [\textit{m}]}} & \multicolumn{1}{l}{\textbf{$\Delta$FDE [\textit{m}]}} & \multicolumn{1}{l}{\textbf{$\Delta$AAE [$^{\circ}$]}} & \multicolumn{1}{l}{\textbf{$\Delta$FAE [$^{\circ}$]}} \\ \hline
CNN-LSTM    & \textbf{0.01} & \textbf{0.02} & \textbf{0.87} & \textbf{0.68} \\
LMU         & 0.04 & 0.07 & 4.02 & 2.92 \\
Transformer & 0.02 & 0.05 & 4.70 & 2.84 \\
GNN         & 0.04 & 0.03 & 2.47 & 2.78                                      
\end{tabular}
\end{table}

\section{Summary}
\label{sec:summary}

This paper evaluated various classic and ML-based forecasting models, including Constant Velocity, Linear neural network, LSTM, LMU, TCNN, GNN (with GAT), and Transformers, in the context of NBA player position forecasting. 

Our benchmarks for determining the optimal input sequence length indicate that a duration of approximately 2\unit{s} is ideal for most architectures. Shorter input windows lack sufficient historical context for accurate predictions, while longer windows yield diminishing returns in predictive accuracy.

To evaluate the effects of unseen trajectories on our models, we benchmark them in a single-team context, i.e., a random split.
For short forecast horizons ($<$ 0.5\unit{s}) all models perform well, with minimal error differences. Constant Velocity and Linear models perform best up to 0.12\unit{s}, thanks to the minimal trajectory changes in this range. At medium forecast horizons (0.12–0.4\unit{s}) Constant Velocity begins to falter due to its inability to capture non-linear motion. Linear and LSTM models start to outperform others. And at long forecast horizons ($>$ 0.4\unit{s}) all models surpass the Constant Velocity baseline. Beyond 1.3\unit{s}, even the Linear model is outperformed by all others, highlighting its limitations in modeling complex temporal dynamics. Especially our CNN-LSTM Model shows the strongest performance across long horizons, benefiting the most from additional context. It reduces FDE by 40\% (from 2.5\unit{m}to 1.5\unit{m}), outperforming pure LSTM models beyond 1.28\unit{s}. Our LMU model achieves slightly higher FDE (1.59\unit{m}) than CNN-LSTM, indicating strong but slightly inferior performance. This suggests potential in combining CNN and LMU using Legendre polynomials. The GNN model performs comparably to LMU (FDE = 1.62\unit{m}) and outperforms the Transformer, likely due to its ability to leverage both attention and graph-context information. And our Transformer model slightly underperforms (FDE = 1.66\unit{m}). We believe this is likely due to insufficient training data, which limits the model’s ability to fully leverage attention mechanisms.

We also assessed the generalizability of our temporal context models (LMU, CNN-LSTM, GNN, Transformer) by testing their performance on unknown (left out) games and players from different teams. 
The CNN-LSTM model offers robust performance (lowest FDE of 1.53\unit{m}), with  marginal increases in error (FDE by 0.02\unit{m}). We observed similar trends for the LMU (0.07\unit{m}), Transformer (0.05\unit{m}), and GNN (0.03\unit{m}) models. So, our models adapt to unfamiliar team dynamics. 

Our findings highlight the importance of integrating context information for accurate and generalizable player position forecasting.
Moreover, our models help mitigate delays inherent in modern signal processing pipelines, offering practical applications in real-time sports analytics and other domains that demand precise motion forecasting.

\bibliography{ION_PLANS}
\bibliographystyle{IEEEtran}

\end{document}